%% file: sample-xelatex.tex
\documentclass[sigconf,screen]{acmart}

\def\BibTeX{{\rm B\kern-.05em{\sc i\kern-.025em b}\kern-.08emT\kern-.1667em\lower.7ex\hbox{E}\kern-.125emX}}

\copyrightyear{2019}
\acmYear{2019}
\setcopyright{acmcopyright}
\acmConference[MM '19]{Proceedings of the 27th ACM International Conference on
Multimedia}{October 21--25, 2019}{Nice, France}
\acmBooktitle{Proceedings of the 27th ACM International Conference on Multimedia (MM '19),
October 21--25, 2019, Nice, France}
\acmPrice{15.00}
\acmDOI{10.1145/3343031.3351044}
\acmISBN{978-1-4503-6889-6/19/10}

\acmSubmissionID{1001}

\usepackage{algorithm}
\usepackage{algorithmicx}
\usepackage{algpseudocode}
\usepackage{mathrsfs}
\usepackage{subfigure}

\begin{document}
\fancyhead{}

%
\title{Adversarial Seeded Sequence Growing for Weakly-Supervised Temporal Action Localization}
\renewcommand{\shorttitle}{Adversarial Seeded Sequence Growing for Action Localization}
%

\author{Chengwei Zhang$^{1+}$,Yunlu Xu$^{2}$,Zhanzhan Cheng$^{23*}$,Yi Niu$^{2}$,Shiliang Pu$^{2}$,Fei Wu$^{3}$,Futai Zou$^{1}$}
\thanks{\textsuperscript{+}Zhang partially did this work during an internship in Hikvision Research Institute.}
\thanks{\textsuperscript{*}Corresponding author.}
\affiliation{
  \textsuperscript{1}\institution{Shanghai Jiao Tong University, Shanghai, China}
}
\email{cwzhang,zoufutai@sjtu.edu.cn}

\affiliation{
  \textsuperscript{2}\institution{Hikvision Research Institute, China}
}
\email{xuyunlu,chengzhanzhan,niuyi,pushiliang@hikvision.com}
\affiliation{
  \textsuperscript{3}\institution{Zhejiang University, Hangzhou, China}
}
\email{wufei@cs.zju.edu.cn}

%
\renewcommand{\shortauthors}{Zhang, Xu and Cheng, et al.}

%

\input{sections/0_abstract.tex}

%
%


%

\maketitle

\input{sections/1_intro.tex}
\input{sections/2_related.tex}

\input{sections/3_methods.tex}
\input{sections/4_exp.tex}

\input{sections/5_conclu.tex}

\section{Acknowledgement}
Chengwei Zhang and Futai Zou were partially supported by National Key Research and Development Program of China (No.2017 YFB0802300), NSFC-Zhejiang Joint Fund for the Integration of Industrialization and Informatization (No. U1509219), and National Key Research and Development Program of China (No. 2018YFB 0803500). THANKS to all members in DAVAR lab.

\bibliographystyle{ACM-Reference-Format}
\bibliography{egbib}


%
\appendix

\end{document}

%% file: sections/0_abstract.tex
\begin{abstract}

Temporal action localization is an important yet challenging research topic due to its various applications.
Since the frame-level or segment-level annotations of untrimmed videos require amounts of labor expenditure,
studies on the weakly-supervised action detection have been springing up.
However, most of existing frameworks rely on Class Activation Sequence (CAS) to localize actions by minimizing the video-level classification loss,
which exploits the most discriminative parts of actions but ignores the minor regions.
In this paper, we propose a novel weakly-supervised framework by adversarial learning of two modules for eliminating such demerits.
Specifically, the first module is designed as a well-designed Seeded Sequence Growing (SSG) Network for progressively extending seed regions (namely the highly reliable regions initialized by a CAS-based framework) to their expected boundaries.
The second module is a specific classifier for mining trivial or incomplete action regions, which is trained on the shared features after erasing the seeded regions activated by SSG.
In this way, a whole network composed of these two modules can be trained in an adversarial manner.
The goal of the adversary is to mine features that are difficult for the action classifier.
That is, erasion from SSG will force the classifier to discover minor or even new action regions on the input feature sequence, and the classifier will drive the seeds to grow, alternately.
At last, we could obtain the action locations and categories from the well-trained SSG and the classifier.
Extensive experiments on two public benchmarks THUMOS'14 and ActivityNet1.3 demonstrate the impressive performance of our proposed method compared with the state-of-the-arts.

\end{abstract}

%% file: sections/1_intro.tex
\section{Introduction}
Temporal action localization, also called action detection, is to localize the temporal locations of actions as well as identify action categories from untrimmed videos, which is a fundamental and challenging problem in video understanding.
Many existing works \cite{Shou2016Temporal,Zhao2017Temporal,Shou2017CDC,Xu2017R,yang2018exploring,Chao2018Rethinking,Lin2018BSN,Alwassel2018ECCV,buch2017sst,Dai2017Temporal,Richard2016Temporal}
make efforts to address this problem in a strong-supervised manner,
where these algorithms rely on fully labeled data (\emph{e.g}. actions annotated with precise starting and ending frames).
However, untrimmed videos are usually very long, so manually annotating action locations usually seems time-consuming and expensive in real applications.
\begin{figure}[!t]
\begin{center}
   \includegraphics[width=0.99\linewidth]{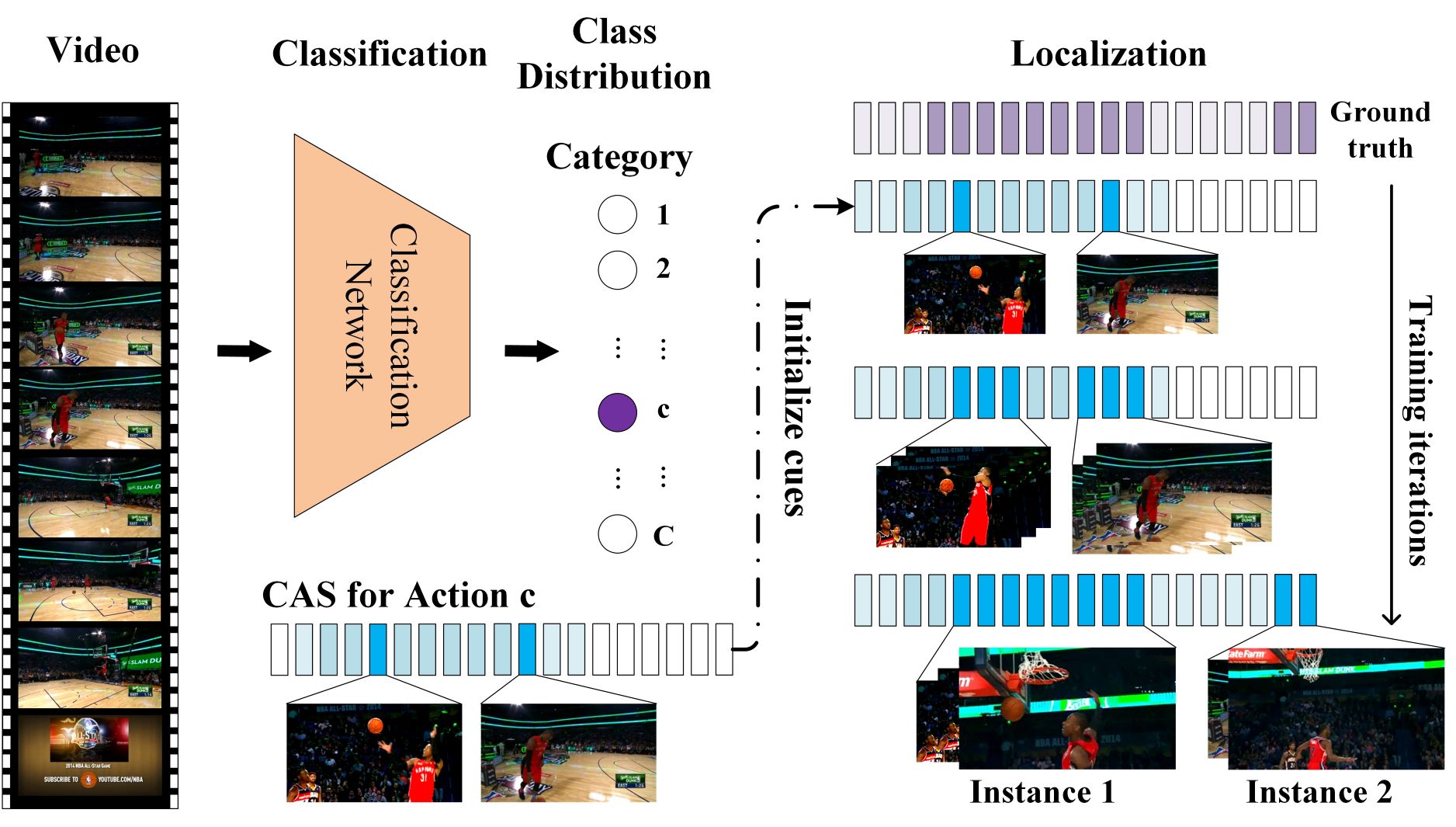}
\end{center}
   \caption{Seed-Grow Mechanism for Action Localization.}
   \label{fig:intro}
\end{figure}

Above issues motivate researchers to weakly-supervised temporal actions detection only using video-level labels (i.e., action categories).
Actually, weakly-supervised temporal action detection is similar to object instance detection in image with only image-level annotations (i.e., object categories).
Inspired by Class Activation Map (CAM) \cite{Zhou2015Learning} used in object detection, STPN \cite{Nguyen2017Weakly} introduces the one-dimension extension of the CAM named as Temporal-CAM, also called Class Activation Sequence (CAS) in recent works \cite{Shou_2018_ECCV}, to locate the temporal activating regions by conducting the classification task with only video-level labels.

Following the CAS-based classification framework, \cite{Shou_2018_ECCV} attempted to design an Outer-Inner-Contrastive(OIC) loss to find salient intervals, and \cite{Paul_2018_ECCV} addressed the co-occurrence problem in action detection for capturing more discriminative patterns.
Recently, work \cite{Xu2018Segregated} focused on the relation learning among actions via an RNN architecture and the CAS mechanism.
All the above CAS-based works are designed for weakly-supervised action localization, and achieve good performance, especially on the evaluation of low IoUs (e.g. IoU=0.1 or 0.2).

However, the CAS-based action detector usually localizes actions in untrimmed videos at the most discriminative action interval, which often appears in action response peak and results in the failure on the evaluation of high localization precision (See Figure \ref{fig:intro}). 
That is, such CAS-based actions detectors tend to fall into two essential issues: 1) Poor performance on the evaluation of long-duration actions due to the peak response problem caused by CAS. For example, only few discrete regions of Instance 1 are activated by CAS, which directly results in the poor results on the long-duration action detection.
2) Missing of trivial or indiscriminative actions as the case of Instance 2 missed in CAS results.

Here, we focus our attention on conquering above issues.
With extensive observations, we find that though CAS tends to generate sparse activating peaks on action regions,
these peaks provide important cues for mining salient parts of actions or indiscriminative actions.
Therefore, an intuitive idea is that to mine more reliable action regions by referring to the estimated action cues, termed as \textit{seeds}. Inspired by the \textit{seed-grow} mechanism in image segmentation tasks \cite{Kolesnikov2016Seed}, we adapt it into the temporal action localization tasks. Differently, we design the following two complementary manners of \textit{grow} in the seeded sequences.
\begin{itemize}
  \item We treat these activated peaks as \textit{seeds} indicating important action cues and then extend the time durations for separated seeds to their boundaries, denoted as \textbf{the first manner of \textit{grow}}.
  \item Simultaneously, we erase the activated peaks from shared feature regions and further conduct a \textit{self-adaptive classifier} for mining potential trivial or indiscriminative actions, denoted as \textbf{the second manner of \textit{grow}.}
\end{itemize}
Above two procedures should be trained in an adversarial manner.
On the one hand, erasing seeded regions of SSG will force the classifier to mine the less discriminative action regions from the feature regions.
On the other hand, the classifier will also drive the seeds to grow, alternately.

In this paper, we propose a new weakly-supervised action detection framework called \textbf{Adversarial Seeded Sequence Growing (ASSG)} by adversarial learning of
a \textbf{Seed Sequence Growing (SSG)} network and a \textbf{self-adaptive action classification} network.
{Specifically,
the \textit{SSG} is responsible for learning independent temporal heatmaps (corresponding to the action occurring probability distribution) for each action category respectively. 
The module takes in the most discriminative regions from CAS as the initial seeds, and progressively expands the seeded regions to neighborhood in a self-guided way.
The \textit{action classifier} devotes to exploiting the trivial missing or incomplete instances. It first erases the high-confidence regions from the SSG and thus has to find new reliable parts by the supervision of video-level class annotations. It worth noting that the module adjusts the training parameters of the shared feature maps with the SSG without any additional learning parameters, so the classifier can further promote the expanding (namely growing) of seeded regions.}
Consequently, these two module are trained in an adversarial manner and jointly contributes to the iteratively growing of reliable regions.
Then the final results, i.e., the action locations and their categories, can be obtained from the well-trained SSG and the classifier.

The main contributions of our paper are summarized as follows:

(1) We propose an end-to-end weakly-supervised action detection approach integrating SSG network and a specific video-level classifier for mining indiscriminative action locations. 
To the best of our knowledge, this is the first work to introduce the \textit{seed-grow} mechanism in temporal action detection.

(2) We train two modules in an \textit{adversarial} manner, which not only helps grow action occurring durations and also mines trivial or indiscriminative actions.

(3) Extensive experiments demonstrate that
our method achieves impressive performance on the challenging  THUMOS'14 \cite{THUMOS14} and ActivityNet1.3 \cite{heilbron2015activitynet:} datasets,
especially on the evaluation of high IoUs which is more valuable than that in low IoUs.

%% file: sections/2_related.tex
\section{Related Work}
\subsection{Temporal Action Localization.}
Temporal action localization aims at identifying the temporal action intervals.
According to the utilized supervision information in model training, we divide existing methods into two categories: the fully-supervised based and the weakly-supervised based.

\textbf{Fully-Supervised Action Detection.}
Most existing works generally train action detection model with frame-wisely action annotations, i.e., each action is annotated with category as well as its starting and ending position.
In early time, sliding windows strategy \cite{oneata2014efficient} with a well-trained action classifier is the traditional solution for temporal detection.
Afterwards, the proposal-based \cite{Shou2016Temporal, Zhao2017Temporal, Xu2017R} methods were developed for effectively {narrowing down the number of candidate instances}.
Specifically,
S-CNN \cite{Shou2016Temporal} used a multi-stage CNN for temporal action localization with extracted robust video feature representation.
SSN \cite{Zhao2017Temporal} applied a watershed temporal actionness grouping algorithm (TAG) \cite{heilbron2015activitynet:} for generating action regions
and then designed the structured temporal pyramid classifiers for identifying actions' categories and their localization. 
Inspired by object detection, SSAD \cite{Lin2017Single}, R-C3D \cite{Xu2017R} and TAL-Net \cite{Chao2018Rethinking} were proposed to detect one-dimensional actions by generalizing 2D spatial proposal mechanism to 1D temporal proposal form.
Recently, a special boundary sensitive network (BSN) \cite{Lin2018BSN} attempted to locate temporal boundaries and further integrated them into action proposals.
In addition, some frame-level segmentation networks \cite{Shou2017CDC,yang2018exploring} were also developed to generate more precise action localizations by conducting the one-dimensional semantic segmentations task.
However, all above works rely on frame-wisely action annotations, which are usually impractical in real applications due to the amounts of labor expenditure. \\

\textbf{Weakly-Supervised Action Detection.}
Recently, action detection with only video-level labels has been studied.
UntrimmedNet \cite{wang2017untrimmednets} introduced an end-to-end model for learning only single-label action categories as well as localizations,
which is the first action detection approach without using frame-wise labels. 
STPN\cite{Nguyen2017Weakly} adopted an attention module to identify a sparse subset of key action segments in a video,
and fused the key segments into action regions via adaptive temporal pooling.
Similarly, AutoLoc \cite{Shou_2018_ECCV} directly learned the boundaries using a novel Outer-Inner-Contrastive (OIC) loss to provide the desired segment-level supervision,
and W-TALC \cite{Paul_2018_ECCV} introduced the Co-Activity Similarity Loss and jointly optimized it with the cross-entropy loss for weakly-supervised detecting temporal actions.
The recent state-of-the-art STAR \cite{Xu2018Segregated} focuses on the relationship learning among multiple actions,
which exploits recurrent networks for assembling expected action instances into high-level feature representation, and then predicted class labels and locations for each action category step-by-step.
Most of them apply the attention mechanism integrating the Class Activation Sequence(CAS) mechanism to capture the most discriminative temporal regions,
while they ignore the long-duration of action occurring problem or the missing of trivial action regions.
Therefore, previous methods usually falls into poor performance on the evaluation of high-IoU localization.
Fortunately, recent work \cite{Zhong2018Step} found the contradiction between classifier and detector and thus designed a step-by-step erasion approach to train the one-by-one classifiers. This provides us the inspiration to further mine the less discriminative features from a complete video iteratively.

\subsection{Weakly-Supervised Object Localization.}
Weakly supervised object localization methods locate target objects using convolutional classification networks.
The widely-used Class Activation Map (CAM) \cite{Zhou2015Learning} can be used to
discover the spatial distribution of discriminative parts,
while they can but find very small and noncontinuous regions of the entire objects.
To handle with the weakness, Hide-and-seek \cite{Singh2017Hide} tried to force the model to see different parts of the image and focused on multiple relevant parts of the object beyond just the most discriminative one. It is implemented by randomly masking different regions of training images in each training epoch. Erasing-based approaches \cite{Wei2017Object} are proposed to mine the complementary object regions other than the former most discriminative parts,
and then use the fused results as the final object localization.
Furthermore, semantic segmentation methods \cite{Kolesnikov2016Seed,huang2018weakly-supervised} treat the salient parts of object as seeds, and then iteratively expand the regions to a definite object boundary. These explorations based on image-level annotations have exemplified the weakly supervised object detection, which could provide valuable experiences to the temporal action localization.

%% file: sections/3_methods.tex
\begin{figure*}[t] \label{fig:overview}
\centering
   \includegraphics[width=0.8\linewidth]{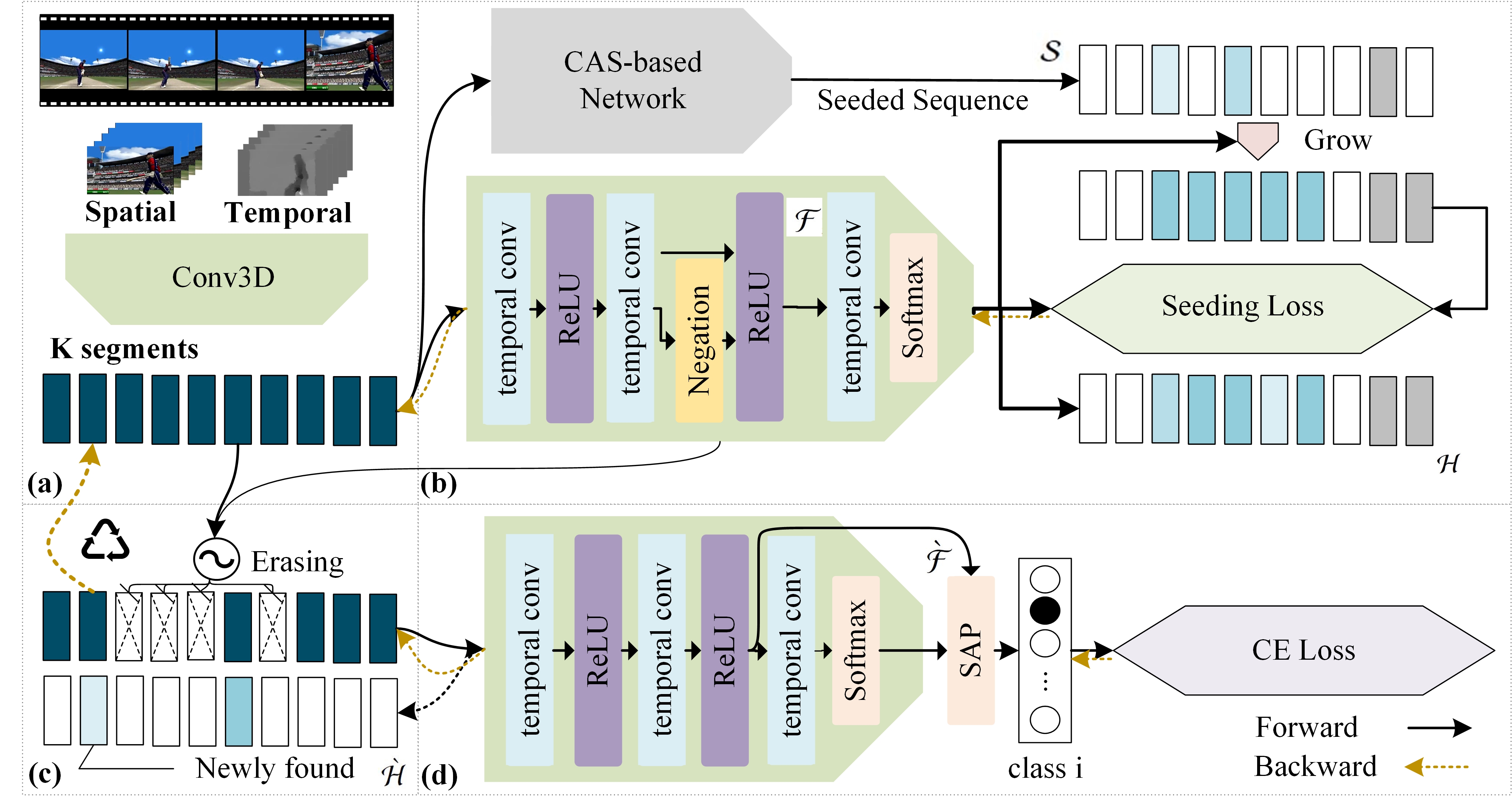}
   \caption{The proposed ASSG architecture. (a) Encoded segmental features from video inputs. (b) The SSG module, with CAS-based results as initial seeds, iteratively extending the high reliable action or background regions using growing rules, which learns the category of reliable input segments by a seeding loss. (c) Erasion of the seeding regions from the shared feature map from SSG. (d) An action classifier, which uses self-adapted pooling (SAP) for feature aggregation into the final class confidence with the cross-entropy loss. }
\label{fig:overview}
\end{figure*}

\section{Methodology}
\subsection{Overview}\label{overview}
Given an untrimmed video, we traditionally trim the sequence into $N$ segments, and encode each segment as a $K$-dimensional feature vector with a pre-trained two-stream video feature extractor.
That is, $X=\{x_t\}_{t=1}^N, x_t \in \mathbb{R}^{K}$.
Our goal is to localize all action instances in videos via an action detection model trained with only video-level annotations.
Here, the annotation of each video is defined as $Y = \{y_i\}_{i=1}^{M}$, $y_i \in \{1, 2, \cdots, \mathcal{C}\}$, where $M$ is the number of categories in this single video and $\{1, 2, \cdots, \mathcal{C}\}$ means the set of action categories.

For this purpose, we design an end-to-end action detection framework composed of two adversarial parts: 1) the Seeded Sequence Growing (SSG) module for extending action occurring durations with pre-fetched initial seeds, which is detailed in Subsection \ref{SSG}, and 2) the self-adaptive action classifier for further exploiting the missing or incomplete instances, which is detailed in Subsection \ref{MSC}.
The two modules are integrated into an entire framework and trained in an adversarial manner, as shown in Figure \ref{fig:overview}.

\subsection{Seeded Sequence Growing} \label{SSG}
The SSG module first uses the reliable weakly-supervised results as initial seeds
to generate reliable supervision of sparse discriminative regions. Then it progressively increases the seeded temporal regions.

\subsubsection{Initial Seeds.}
\textit{Foreground seeds} can be generated from CAS-based networks \cite{Nguyen2017Weakly, Shou_2018_ECCV, Xu2018Segregated}.
The sparse and high reliable action regions are the peaks in the CAS activations by a relatively high threshold.

Background refers to those non-action occurring durations.
Assuming that a background region is likely to appear between two action durations when it comes to a scene change,
while action occurring at a time always has the consistent shot motion,
we utilize the saliency detection \cite{wang2017untrimmednets} to capture shot changes as the probable \textit{background seeds}.

Here, we formalize the initializing seeds as $\mathcal{S}=\{\mathcal{S}_c\}_{c=0}^C$, 
where $c=0$ represents the background, and $c \in \{1, \cdots, \mathcal{C}\}$ represents each action category respectively.

\subsubsection{Backbone Network.} \label{SSGN}
The SSG network is {learning independent temporal heatmaps for each action category. It is designed to progressively label more reliable action locations from the initial seeds $\mathcal{S}$. 

Concretely,
the backbone of SSG module first stacks two temporal convolutional layers (striding filters along time dimension), in which each temporal convolutional layer follows the setting as \{filters=512, kernel size=1 , stride=1\}.
A ReLU layer follows each temporal convolution.
On top of the SSG is also a temporal convolution layer but for producing the {\emph{class heatmaps} for each temporal segments  $\mathcal{H} = \{\mathcal{H}_{c,t} | c\in \{0, \cdots, C\}, t\in \{1, \cdots, N\}\}$, in which $\mathcal{H}_{c,t}$ means the class $c$ probability distribution of the $t$-th segment in the video.

\subsubsection{Growing Strategy.}
Inspired by the growing strategy of dynamic supervision \cite{huang2018weakly-supervised}, we propose a one-dimension growing policy to dynamically enlarge the seeded sequences on each independent class heatmap $\mathcal{H}_{c,t}$.
Once initialized, the current action regions are grown by expanding those seeds $\mathcal{S}$ to neighboring unlabeled locations $\mathcal{N}(\mathcal{S})$ via the growing criterion $\mathcal{G}$:
\begin{equation}
\mathcal{G}(\mathcal{H}_{c,t},\mathcal{S}_c,\theta_{g}^{c})=
\begin{cases}
True, &l \in \mathcal{N}(\mathcal{S}_c), ~\mathcal{H}_{c,t}\geq\theta_{g}^{c} ~ \\
&\text{and}~ c=\mathop{\arg\max\limits_{c'}}\mathcal{H}_{c',t}\\
False, &otherwise.
\end{cases},
\end{equation}
where $\theta_{g}$ is the pre-set \emph{growing threshold} value respectively for each action class and the background. Here, we use a simple definition of $\mathcal{N}(\mathcal{S})$ to be the set of locations next to each seed in $\mathcal{S}$.
Considering this criterion, if $\mathcal{G}$ is true, we label the category of the $t$-th segment with class $c$ as newly added supervision regions.
{Iteratively, we motivate the activations of both the action occurring and the background durations on heatmaps alternately with the grown supervision.}

In practice, 
{since the \textit{co-occurrence} locations can not be assigned to two different classes when conducting the \textit{seed-grow} mechanism in original temporal segmentation framework,
we generate separated seeds for each action class (including the background) and expand the seeded regions respectively.
That is, the SSG predicts individual action occurring regions one-by-one with growing policy for each class.
}

\subsection{Self-Adaptive Action Classifier} \label{MSC}
This module is designed to mine the relatively long or trivial actions, which shares the feature map with SSG.
It first erases the most discriminative part dynamically activated by SSG, and then predicts the action class by directly \emph{aggregating} the output of shared feature maps into classification confidence scores.
In this way, the classifier adaptively updates the shared parameters supervised by video class annotations without adding extra parameters. 

To be specific, in the first step, we \emph{extract} the foreground feature maps from the entire maps $\mathcal{F}$ in SSG, and then \emph{erase} the highly activated regions to generate the remaining feature maps $\grave{\mathcal{F}}$.
As $\mathcal{F}$ contains mixed activations of the foreground and the background, we need to draw only the foreground features for classification. Therefore, a pair of opposite ReLU activations is designed to generate $\mathcal{F}$, which forces the foreground or background seeds to grow in the positive or negative activation parts respectively.
Naturally, the classifier can obtain the foreground features by a ReLU layer. The erasion is simply implemented by thresholding on the activation values.

The next step is the \emph{aggregation} of feature maps.
The common \emph{aggregating} approach like global max-poling (GMP) \cite{oneata2014efficient}
or global average-pooling (GAP) \cite{Zhou2015Learning} 
is not suitable for this task, since the former ignores too many less discriminative regions and in the later case, the global feature will be overwhelmed by the large-scale occupied background segments.
For the purpose of inspiring full potential of the classifier, 
we design a \textbf{Self-Adaptive Pooling (SAP)} approach for straightforward video class prediction.
The SAP is to re-balance the weights of segments with an off-the-shelf attention weights $A_{c,t}(X)$ for class $c$ at temporal location $t$, which could be achieved from the learned feature maps shared with the SSG as

\begin{equation}\begin{aligned}
SAP(X) &= \sum_{t=1}^{N}A_{\ast,t}(X)\cdot \grave{\mathcal{H}}_{\ast,t}. \\
\end{aligned}
\end{equation}

Here, $\cdot$ is the dot product operation of two scalars.
Similarly to $\mathcal{H}$, $\grave{\mathcal{H}}$ here is the activation distribution after erasion. The self-adaptive weighted aggregation of $\grave{\mathcal{H}}$ over the entire $N$ temporal segments results in the entire video-level class distribution.}

In the equation, the attention weights $A_{c,t}(X)$ can be formulated by
\begin{equation}
A_{c,t}(X) = \frac{e^{(\sum_{i=1}^{|f_{c,t}(X)|}f_{c,t}^i(X))}}{\sum_{t=1}^{N}  e^{(\sum_{i=1}^{|f_{c,t}(X)|}f_{c,t}^i(X))}},
\end{equation}
where {$f(\cdot)$ represents the mapping functions for the foreground features $\grave{\mathcal{F}}$ from the network inputs. $|f_{c,t}(X)|$ represents the feature dimensions at each location on $\mathcal{F}$.

Note that, assuming the highly activated regions are likely to be the interested action occurrences, $A(X)$ is the self-adaptive weight directly computed from the feature map without any extra explicit attention modules, .

\subsection{Training of ASSG} \label{loss}
The two modules (i.e.,the SSG and the action classifier) are integrated into a whole network and trained in an adversarial manner.

\subsubsection{Seeded Sequence Growing Loss.}
$T_{c}$ is a set of temporal locations that are identified as action class $c$. 
Here, the seeding loss $L_{seed}$ is defined as:
\begin{equation} \label{seedloss}
\begin{split}
L_{seed}=&-\frac{1}{\sum_{c\in{[0,C]}}|T_c|}\sum_{c\in{[0,C]}}\sum_{t\in{T_c}}log\mathcal{H}_{c,t} \\  
\end{split}
\end{equation}
The SSG learns the parameter by optimizing the seeding loss function $L_{seed}$,
which encourages the networks to match reliable localization cues $T_{c}$, including the foreground ($c\in{[1,C]}$) and the background ($c=0$),  but is agnostic about the rest of the locations.

\subsubsection{Action Classification Loss.}
The \textit{classification loss} is defined as the cross-entropy loss over multiple categories by
\begin{equation} \label{clsloss}
\begin{split}
L_{class}&=
         -\sum_{c=1}^C\hat{y}_c\log{SAP(\hat{y}_c|X)},\\
\end{split}
\end{equation}
where $\hat{y}_c$ represents the ground truth of the action category.

\subsubsection{End-to-end Training.}
The whole network is trained in an adversarial manner.
By erasing the seeded regions activated by SSG, the classifier branch poses a more difficult task to discover minor or even new action regions.
Alternately, the classifier will also boost the seeds growing and generate more reliable regions.
The training procedures are illustrated in Algorithm 1.
\begin{algorithm}[htb]
\caption{Framework of \textbf{ASSG}.}
\label{alg:Framwork}
\begin{algorithmic}[1]
\Require
    Training data, $\mathcal{V} =\{\mathcal{X}, y_c\}$; growing threshold, $\theta_g$; adversarial threshold $\theta_a$; sequence length $N$; total categories $C$;
	initial seeds $\mathcal{S}$
\Ensure
	Enhanced CAS heatmaps $\mathcal{H}$; prediction label $c$;
	
\State Initialize $\mathcal{H}$;
\State Initialize the shared deep feature map $\mathcal{F}$; 
\While{training not converge}
\For{$t = 0 \to N-1$}
\For{$c = 0 \to C$}
\If{$\mathcal{G}(\mathcal{H}_{c,t}(\mathcal{X}),\mathcal{S},\theta_{g}^{c})$}
\State  the location at t-th segment is labeled as c;
\Else
\State the location at t-th segment keeps unlabeled state;
\EndIf
\EndFor
\EndFor	

\State Update $\mathcal{H}$ with the seeding loss $L_{seed}$.
\State Obtain high reliable regions: $U_{c}$  = \{$t|\mathcal{H}_{c,t} > \theta_a$ \};
\State Erase clips at temporal location in sets $\{U_{c}\}_{c=1}^C$ for each category from ${\mathcal{F}}$ to obtain $\grave{\mathcal{F}}$;
\State Calculate SAP($\mathcal{X}$);
\State Update $\grave{\mathcal{F}}$,$\mathcal{F}$ (sharing parameters) with the cross-entropy loss $L_{class}$.
\State Compute updated $\mathcal{H}$;
\EndWhile

\end{algorithmic}
\end{algorithm}

\subsection{Location Prediction}

We use the predicted heatmap by the SSG module to generate temporal action proposals.
{As $\mathcal{H}_{c,t}$ denoted in Subsection \ref{SSGN} indicates the probability in the class heatmaps.}
Here, we fuse the separately trained two-stream network predictions.
For each class \textit{c} in the corresponding heatmap, each proposal [$N_{start},N_{end}$] is assigned to a score by:
\begin{equation} \label{twostream}
\sum_{t=N_{start}}^{N_{end}}
\frac{ [\lambda\cdot \mathcal{H}_{c,t}^{RGB} + (1-\lambda)\cdot \mathcal{H}_{c,t}^{flow}]}
{N_{end}-N_{start}+1}
\end{equation}
in which we fuse the probability values of RGB and optical flow streams by the modality ratio $\lambda$.
For final detection, we perform non-maximum suppression (NMS) among temporal proposals of each class by removing highly overlapped ones.

%% file: sections/4_exp.tex
\section{Experiments}
\subsection{Datasets and Evaluation}
\textbf{Datasets.} \textbf{THUMOS'14 \cite{THUMOS14}} is a popular dataset for action localization task, which consists of 20 action classes, and only its validation and testing set contains temporal annotations. Since the training set contains no temporal annotations, the fully-supervised algorithms use the 212 untrimmed videos in validation set to train the network and the 200 testing videos to evaluate the algorithm. To facilitate comparisons, we follow this conventional protocol but without using the temporal annotations for training. \textbf{ActivityNet v1.3 \cite{heilbron2015activitynet:}} covers 200 activity classes. We also use the 10,024 training set videos without temporal annotations for training and 4926 validation videos for validating. In this section, we report our results on both THUMOS'14 and ActivityNet v1.3.\\

\textbf{Evalution Metrics.} In action localization task, mean Average Precision (mAP) is used as evaluation metrics, where Average Precision (AP) is calculated on each action class when the prediction is classified correctly and its temporal overlap Intersection over Union (IoU) with the ground truth segments exceeding the evaluation threshold.
The ablation study are performed on the THUMOS’14 dataset and evaluated with average mAP (Ave-mAP) by calculating the multiple overlap IoU with thresholds varying from 0.1 to 0.5.
The overall performance compared with the state-of-the-arts is evaluated with average mAP from 0.1 to 0.5 on THUMOS'14 and 0.5 to 0.95 on Activitynet v1.3.

\subsection{Implementation Details}
We implement our algorithm using Caffe \cite{Jia2014Caffe}. For comparison, we employ the common procedures described in \cite{Nguyen2017Weakly, Xu2018Segregated} to uniformly sample 400 segments from  each video. For extracting visual features, then we use the two-stream I3D network described in  \cite{carreira2017quo} pre-trained on Kinetics dataset \cite{zisserman2017the}. For the CAS-based network, we realize the ST-GradCAM with the pre-defined parameters described in \cite{Xu2018Segregated} as our default setting. Note that we train the ST-GradCAM without a specific sub-module for repetition alignment, which needs additional annotations of action frequency. Our proposed network is trained by Adam optimizer with initial learning rate $10^{-4}$ on both streams. For the growing threshold $\theta_{g}$, we set both foreground and background threshold as 0.99. Besides, the earsing threshold $\theta_a$ is set to 0.4 on both datasets. In location prediction, we empirically set $\lambda$ to 0.3. \\

\begin{figure*}[t] \label{fig:sample}
\centering
   \includegraphics[width=0.9\linewidth]{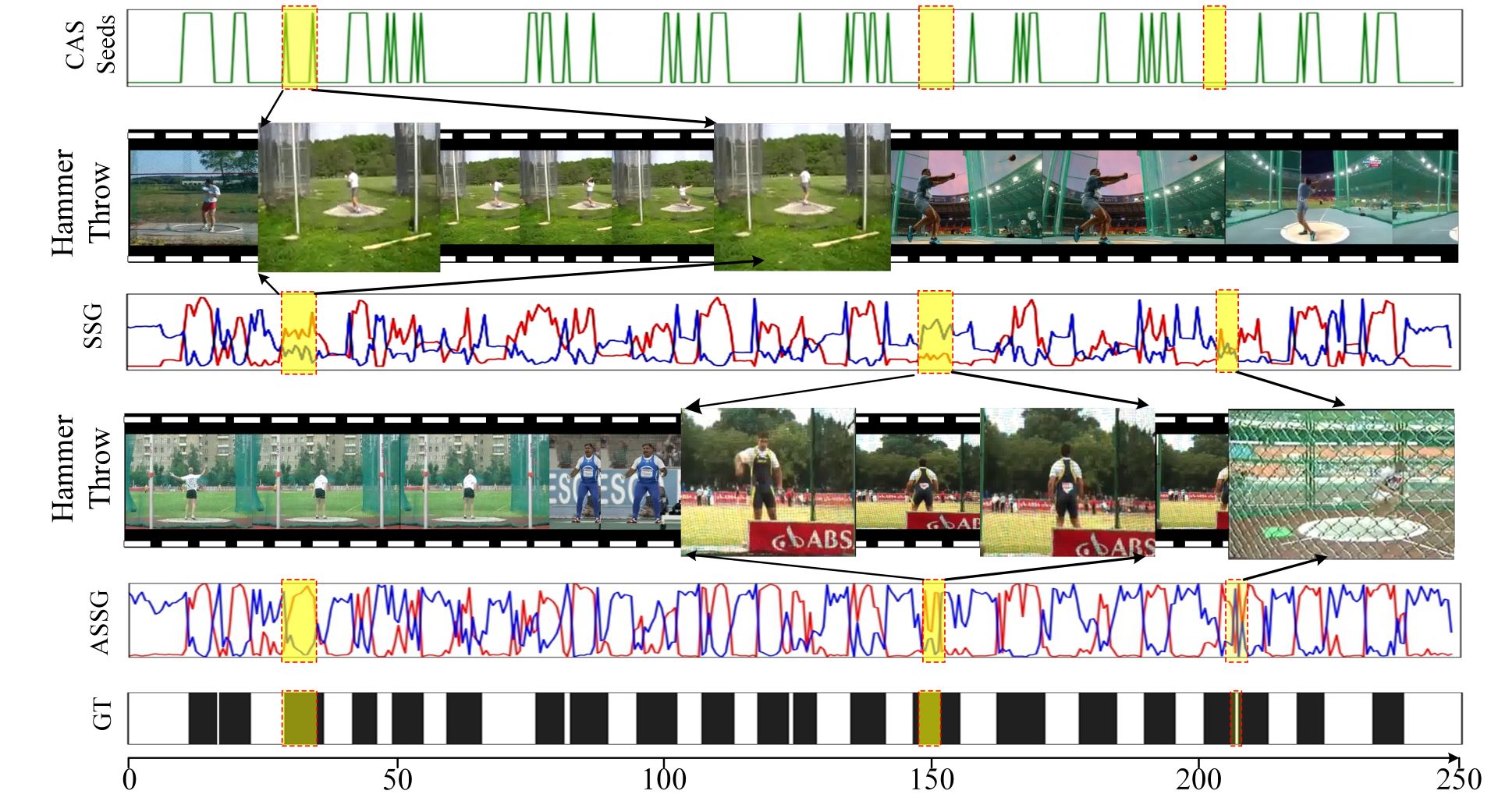}
   \caption{Visualization of action localization by ASSG network. Temporal confidence distribution (the predicted heatmap in SSG) of the action class \textit{Hanmmer Throw} is denoted in red curve and the prediction of the \textit{background} is in blue curve. The \textit{yellow} locations are selected samples of enhanced detection compared with the common CAS results.}
\label{fig:sample}
\end{figure*}

\begin{figure}[h]
  \centering
  \subfigure[The growing threshold $\theta_g$]{
    \label{fig:subfig:a}
    \includegraphics[width=.48\columnwidth]{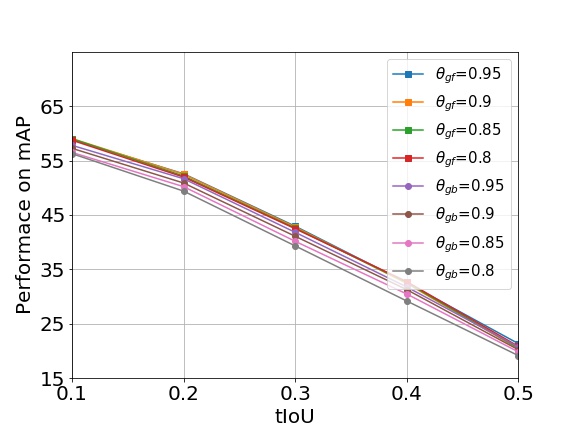}}
  \subfigure[The erasing threshold $\theta_a$]{
    \label{fig:subfig:b}
    \includegraphics[width=.48\columnwidth]{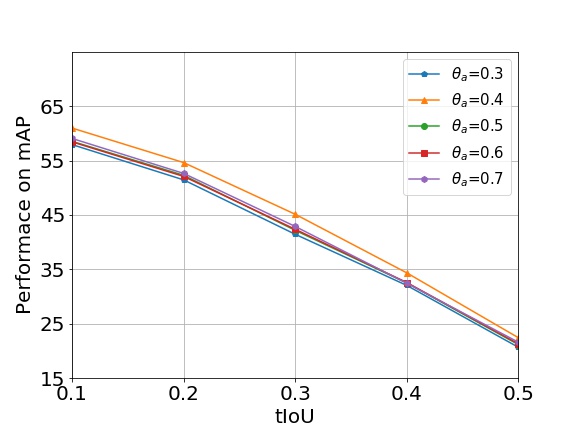}}
  \caption{Performance with different thresholds values. Figure (a) shows mAP evaluation of different $\theta_g$ choices, where  $\theta_{gf}$ and $\theta_{gb}$ means the foreground and background threshold respectively. Figure (b) shows mAP evaluation of different $\theta_a$ choices.}
  \label{fig:threshold}
\end{figure}

\subsection{Ablation Study}
We conduct ablation study on two pre-defined thresholds, effect of feature aggregation methods and different modules. Qualitative evaluation is also provided in this section.

\subsubsection{Thresholds.}
We pre-define two thresholds in our ASSG framework: $\theta_g$ for deciding to label the segment or not in the seeded sequence growing rule, and $\theta_a$ for erasing the high activated regions from SSG in the classification branch, respectively.

For the growing threshold $\theta_{g}$, we could set different thresholds for different classes and background respectively. For convenience, we set the same threshold number for all the actions as foreground threshold $\theta_{gf}$ and another background threshold  $\theta_{gb}$.
Then we fix the $\theta_{gb} = 0.99$ and vary the $\theta_{gf}$ from 0.8 to 0.95 with a step size 0.05, vice verse.
As shown in Figure \ref{fig:subfig:a}, the choice of different threshold values has small influence on the final results, which is convenient and robust for network training.

For the adversarial threshold $\theta_{a}$, it strikes a balance between the adversarial training of two modules. We set the threshold from 0.3 to 0.7 and observe that the localization performance is boosted when the threshold $\theta_a$ = 0.4 as shown in Figure \ref{fig:subfig:b}. Larger value or smaller would introduce more background noises and thus slightly influence the minor region mining.\\ 

\subsubsection{Feature Aggregation.}
In the self-adaptive action classifier module, we aggregate the shared features directly into classification scores without any additional learning parameters. Instead of the common GMP and GAP approach, we introduce an aggregation method called SAP in Subsection \ref{MSC}. Here we discuss the different aggregation types used in our classifier in Table \ref{ablation:SAP}.

We conclude that the low performance of GAP is mainly because the foreground features are overwhelmed by the background segments, while GMP ignores too many regions.
Our proposed SAP achieves best performance at all IoUs compared with the existing GMP and GAP, which verifies the effectiveness of our well-designed self-adaptive weighted aggregation method SAP.\\

\begin{table}[h]
	\begin{center}
		\setlength{\tabcolsep}{1.8mm}
		\caption{Evaluation of different aggregation methods in terms of mAP@IoU on THUMOS'14.
		}
		\scalebox{1}{
			\begin{tabular}{cccccc}
			 \toprule
			  Methods& 0.1  & 0.2  & 0.3  & 0.4  & 0.5   \\
			  \midrule
		      GMP & 52.8 & 46.4 & 37.6 & 29.0 & 18.4  \\
		      GAP & 50.6 & 45.1 & 36.5 & 27.9 & 17.5 \\
		      SAP & \textbf{60.1} & \textbf{54.6} & \textbf{45.1} & \textbf{34.3} & \textbf{22.4} \\
			  \bottomrule
			\end{tabular}
			\label{ablation:SAP}
		}
     \end{center}
\end{table}	

\subsubsection{Architecture and Modules.}
We study the effects of different modules in the entire framework. The CAS, CAS w/ SSG (short in SSG), and the CAS w/ SSG w/ classifier (short in ASSG) results are shown in Table \ref{ablation:2}.
The SSG module boosts the evaluation of the top reported CAS average mAP \cite{Xu2018Segregated} from 24.4\% to 34.2\%. And the additional classification branch for adversarial training jointly denoted as ASSG further increases the mAP value of SSG by 9.3\%, definitely a large margin. Each of the two modules plays an important role in improving the detection results.\\

\begin{table}[h]
	\centering
	\caption{Evaluation of different modules in ASSG on THUMOS'14.}
\resizebox{0.9\columnwidth}{!}{
	\begin{tabular}{ccccc}
		\toprule
		Reported CAS  \cite{Xu2018Segregated}   &  $\surd$         & $\surd$ &  $\surd$ \\ 
		SSG (CAS w/ SSG)    &          & $\surd$ &  $\surd$   \\
		ASSG (CAS w/ SSG  w/ classifier)   &          &          &  $\surd$ \\
		\midrule
		Ave-mAP(\%)      &  24.4    & 34.2 & \textbf{43.5} \\
        \bottomrule		
	\end{tabular}
}
	\label{ablation:2}
\end{table}

\subsubsection{Qualitative Evaluation.}
Visualization results are shown in Figure \ref{fig:sample}. We randomly select a test video in THUMOS'14 dataset (including the action \textit{Hammer Throw}) for the trained proposed ASSG network initialized with the reimplemented CAS baselines \cite{Xu2018Segregated}. For further demonstrating the performance of ASSG, we qualitatively analyze the effective of the \textit{growing} mechanism in different parts of the entire network.
\begin{itemize}
  \item The \textit{seeds} are thresholded results from reimplemented top CAS-based network \cite{Xu2018Segregated}, which effectively detects the most discriminative parts in the videos by a recognition network and fails in the evaluation of high quality detection shown as the missing regional or entire action durations highlighted in \textit{yellow} on top line. For instance, in the first region denoted in yellow rectangles, only the start and end point are activated and in the later two yellow-filled parts, the entire action instances are missing.
  \item We can obviously find the SSG detections expand to some less discriminative parts, which could lead to better detection proposals than the direct CAS result. The improvement verifies the effectiveness of seed-grow mechanism introduced into the SSG for temporal action detection.
  \item Finally, the entire proposed network \textit{ASSG} leads to a more satisfying localization results as the action and complementary background regions interconnect each other tightly. When compared with the single SSG module, we also find that the independent prediction of each foreground class and the background respectively in ASSG holds more confidence (i.e. the prediction probabilities are much higher than the single SSG). We attribute the advantage to the adversarial training of the two modules, which makes it more difficult for the classifier to identify the video class and thus effectively motivates it to mine more minor regions and to expand the small seeded region to its precise boundaries.
\end{itemize}

\subsection{State-of-the-Art Comparisons}
We compare our model with the state-of-the-art weakly-supervised and fully-supervised methods on THUMOS'14 and ActivityNet1.3 benchmarks.
Table \ref{comparison:thumos} and Table \ref{comparison:anet} summarize the results.

\begin{table}[b!]
	\begin{center}
		\setlength{\tabcolsep}{1.5mm}
		\caption{Comparison with state-of-the-arts on THUMOS'14.
		}
		\scalebox{0.8}{
			\begin{tabular}{ccccccccc}
				\toprule
				Method &  Label & 0.1  & 0.2  & 0.3  & 0.4  & 0.5 & 0.6 & 0.7  \\
				\midrule
				Richard et al. \cite{Richard2016Temporal} & strong &39.7& 35.7 & 30.0 &23.2 &15.2 & --   & -- \\
				S-CNN \cite{Shou2016Temporal}&strong&47.7& 43.5 & 36.3 &28.7 &19.0 & 10.3 & 5.3 \\
				CDC \cite{Shou2017CDC} &strong   & -- &  --  & 40.1 &29.4 &23.3 & 13.1 &7.9\\
				Gao et al. \cite{Gao2017Cascaded}  &strong  &54.0& 50.9 & 44.1 &34.9 &25.6 & 19.1 &9.9 \\
				Xu et al. \cite{Xu2017R}  &strong   &54.5& 51.5 & 44.8 &35.6 &28.9 & --   & --\\
				SSN \cite{Zhao2017Temporal}   &strong &\textbf{66.0}& \textbf{59.4} & 51.9 &41.0 &29.8 & 19.6   & 10.7\\
				SSAD \cite{Lin2017Single}  &strong   & 50.1 & 47.8 & 43.0 & 35.0 & 24.6 & --   & --\\
				TPC \cite{yang2018exploring}   &strong   & -- & -- & 44.1 & 37.1 & 28.2 & 20.6   & 12.7 \\
				TALNet \cite{Chao2018Rethinking}  &strong   &59.8& 57.1 & 53.2 &\textbf{48.5} &\textbf{42.8} & --   & --\\
				Alwasssel et al.\cite{Alwassel2018ECCV}&strong  &49.6& 44.3 & 38.1 &28.4 &19.8 & --   & --\\
				BSN \cite{Lin2018BSN}   &strong         &--  & --   & \textbf{53.5} &45.0 &36.9 & \textbf{33.8} &\textbf{20.8}\\ 
				\midrule
				UntrimmedNet \cite{wang2017untrimmednets}  & weak &44.4& 37.7 & 28.2 &21.1 &13.7 & --   & -- \\
				Hide-and-Seek \cite{Singh2017Hide}&weak                   & 36.4& 27.8 & 19.5 &12.7 &6.8 & --   & -- \\
			    Zhong et al. \cite{Zhong2018Step}  & weak  & 45.8 &39.0 & 31.1& 22.5 &15.9 & -- & -- \\
				AutoLoc \cite{Shou_2018_ECCV} & weak  & -- & -- & 35.8 & 29.0 &21.2 & \textbf{13.4} & 5.8 \\
				W-TALC \cite{Paul_2018_ECCV}  & weak & 55.2 & 49.6 & 40.1 & 31.1 & 22.8 & --  & \textbf{7.6}\\
				STPN \cite{Nguyen2017Weakly} & weak &52.0& 44.7 & 35.5 &25.8 &16.9 & 9.9   & 4.3 \\
				STAR \cite{Xu2018Segregated} & weak & \textbf{68.8}& \textbf{60.0} & \textbf{48.7} & \textbf{34.7}& \textbf{23.0} & --  & -- \\
				\midrule
				STPN-CAS w/ ASSG &weak& 55.6 &	49.5 & 41.1 & 31.5 & 20.9 & 13.7 & 5.9 \\
				STAR-CAS w/ ASSG &weak& 65.6 &	59.4 & \textbf{50.4} & \textbf{38.7} & \textbf{25.4} & \textbf{15.0} & 6.6 \\
				\bottomrule
			\end{tabular}
			\label{comparison:thumos}
		}
	\end{center}
\end{table}	
\textbf{Results on THUMOS'14.}
The comparison results between the proposed ASSG and other state-of-the-art models are shown in Table \ref{comparison:thumos}.
As ASSG is a framework enhancing the CAS detections, its seeds can be initialized by various CAS-based models. In this perspective, we conduct ASSG based on two typical structure of CAS-based networks, respectively with CAS results from STPN \cite{Nguyen2017Weakly} as initial seeds (denoted as \textit{STPN-CAS w/ ASSG}) and CAS results from a constrained STAR \cite{Xu2018Segregated} (excluding the specific sub-module with additional action frequency annotation) as initial seeds (denoted as \textit{STAR-CAS w/ ASSG}).
For a fair comparison, we also follow their prediction operations to fuse attention weights with the CAS by generating proposals separately. 

We find that STPN-CAS w/ ASSG improves the performance by a large margin compared to the original STPN result, similarly in STAR-CAS w/ ASSG compared with STAR.
The STAR-CAS w/ ASSG outperforms all other \textit{weakly-supervised} methods with multiple overlap IoU thresholds varied from 0.3 to 0.6.
For instance, when the IoU threshold used in evaluation is set to 0.5, ASSG network improves the state-of-the-art results from 23.0\% to 25.4\%.
It is noting that at IoU threshold of 0.1 and 0.2, our ASSG still achieves impressive performance, which surpasses all other methods except for STAR \cite{Xu2018Segregated}. While STAR used additional action frequency annotations (different from all existed weakly-supervised works), and it also reported results without frequency annotations, i.e., avg-mAP ranging from 0.1 to 0.5 is 44.0\%, which falls behind our ASSG 47.9\% by a 3.9\%.
In specific, we attribute the effectiveness of our approach above CAS-based results \cite{Nguyen2017Weakly,Shou_2018_ECCV,Paul_2018_ECCV, Xu2018Segregated} (especially in higher IoU thresholds) to the ability of ASSG to mine less discriminative action regions, which results in more precise boundaries and completeness of detection.

We also compare the results with the \textit{fully-supervised} methods. The performance of our weakly-supervised model with only video-level class annotations in training, even achieves comparable results with the state-of-the-art strong-supervised ones, by only 0.4\% behind the TALNet \cite{Chao2018Rethinking} at IoU threshold of 0.1 and 3.1\% behind the BSN \cite{Lin2018BSN} at IoU threshold of 0.3. \\


\textbf{Results on ActivityNet1.3.}
Table \ref{comparison:anet} shows the results on ActivityNet1.3 dataset. Note that, the dataset characteristics differ largely from those in THUMOS'14 that many videos in ActivityNet1.3 are including relatively long action durations per instance. Therefore, THUMOS'14 is a better dataset for evaluating most action localization methods, which is also claimed in \cite{Chao2018Rethinking}. Since our designed framework is to expand the action regions from the initial small and sparse reliable regions, connecting all the discrete parts into a unified long-duration action poses a great challenge. 

Results show that ASSG gets better overall performance than all the existing weakly-supervised results by increasing the mAP at IoU threshold of 0.5 and 0.75 from 31.1\% to 32.3\% and 18.8\% to 20.1\% respectively. We do not care about the mAP at IoU threshold of 0.95 since the precision is relatively low, which makes little sense in current situations. Although the performance gain is smaller compared to that in THUMOS'14, the improvement performance also verifies the common effectiveness on both datasets.

\begin{table}[h]
	\begin{center}
		\setlength{\tabcolsep}{1.5mm}
		\caption{Comparison with state-of-the-arts on ActivityNet1.3.
		}
		\scalebox{0.9}{
			\begin{tabular}{ccccc}
			 \toprule
			 Method &  Label & 0.5  & 0.75  & 0.95  \\
			 \midrule
			  Singh et al. \cite{Singh2016Untrimmed} & strong  & 34.5 & --   & -- \\
			  CDC \cite{Shou2017CDC} &strong  & 45.3 & 26.0       & 0.2     \\
			  SSN \cite{Xiong2017A} &strong  & 39.1 & 23.5      & 5.5     \\
			  SSAD \cite{Lin2017Single}  &strong  & 49.0 & 32.9 &  7.9    \\
			  Chao et al. \cite{Chao2018Rethinking} &strong  & 38.2 & 18.3       & 1.3     \\
			  BSN \cite{Lin2018BSN}  &strong   & \textbf{52.5}& \textbf{33.5}& \textbf{8.9}  \\
			 \midrule
		      STPN \cite{Nguyen2017Weakly} & weak & 29.3  &  16.9     & 2.6  \\
			  STAR \cite{Xu2018Segregated} & weak & \textbf{31.1} & \textbf{18.8}  & \textbf{4.7} \\%
			 \midrule
		      STAR-CAS w/ ASSG &weak& \textbf{32.3} & \textbf{20.1} & 4.0   \\
			 \bottomrule
			\end{tabular}
			\label{comparison:anet}
		}
     \end{center}
\end{table}

%% file: sections/5_conclu.tex
\section{Conclusion}
By observing the weakness of CAS-based approach that only the most discriminative parts can be detected,
we extend the \textit{seed-grow} mechanism to our weakly-supervised temporal action detection.
We design a framework of two modules, an SSG and an action classifier respectively,
which jointly help small and sparse action occurring durations grow.
The two modules are trained in an adversarial manner.
The operation of erasing seeded regions forces the classifier to handle with a more difficult task by focusing on less discriminative regions.
Alternately,
the classifier drives the seeds to grow. 
Extensive experiments demonstrate that
our ASSG achieves superior performance on the challenging THUMOS'14 above all other weakly-supervised methods, especially on the evaluation of high IoUs,
and has impressive results on the ActivityNet1.3 datasets as well.